\documentclass[sigconf]{acmart}

\setcopyright{acmcopyright}
\copyrightyear{2024}
\acmYear{2024}

\usepackage{amsmath,amsfonts}
\usepackage{algorithmic}
\usepackage{graphicx}
\usepackage{textcomp}
\usepackage{xcolor}
\usepackage{algorithm}
\usepackage{subfigure}
\usepackage{diagbox}
\usepackage{threeparttable}
\usepackage{multirow}
\usepackage{color}
\usepackage{bbold}
\usepackage{graphicx}
\usepackage{makecell}
\usepackage{caption}
\usepackage{boldline}
\usepackage{colortbl}
\usepackage{wrapfig}
\usepackage{hyperref}
\usepackage{adjustbox}

\begin{document}

\title{Fairness in Large Language Models: A Taxonomic Survey}

\author{Zhibo Chu}
 \email{zb.chu2001@gmail.com}
\affiliation{%
   \institution{Florida International University}
   \city{Miami, FL}
   \country{USA}
}

\author{Zichong Wang}
 \email{ziwang@fiu.edu}
\affiliation{%
   \institution{Florida International University}
   \city{Miami, FL}
   \country{USA}
}

\author{Wenbin Zhang}
\authornote{Corresponding author.}
 \email{wenbin.zhang@fiu.edu}
\affiliation{%
   \institution{Florida International University}
   \city{Miami, FL}
   \country{USA}
}


\begin{abstract}
Large Language Models (LLMs) have demonstrated remarkable success across various domains. However, despite their promising performance in numerous real-world applications, most of these algorithms lack fairness considerations. Consequently, they may lead to discriminatory outcomes against certain communities, particularly marginalized populations, prompting extensive study in fair LLMs. On the other hand, fairness in LLMs, in contrast to fairness in traditional machine learning, entails exclusive backgrounds, taxonomies, and fulfillment techniques. To this end, this survey presents a comprehensive overview of recent advances in the existing literature concerning fair LLMs. Specifically, a brief introduction to LLMs is provided, followed by an analysis of factors contributing to bias in LLMs. Additionally, the concept of fairness in LLMs is discussed categorically, summarizing metrics for evaluating bias in LLMs and existing algorithms for promoting fairness. Furthermore, resources for evaluating bias in LLMs, including toolkits and datasets, are summarized. Finally, existing research challenges and open questions are discussed.  
\end{abstract}


\maketitle


\section{Introduction}
\label{sec:introduction}

\sloppy

Large language models (LLMs) have demonstrated remarkable capabilities in addressing problems across diverse domains, ranging from chatbots~\cite{glaese2022improving} to medical diagnoses~\cite{wang2023chatcad} and financial advisory~\cite{shahfinaid}. Notably, their impact extends beyond fields directly associated with language processing, such as translation~\cite{yao2023empowering} and text sentiment analysis~\cite{nasukawa2003sentiment}. LLMs also prove invaluable in broader applications including legal aid~\cite{yu2022legal}, healthcare~\cite{singhal2023large}, and drug discovery~\cite{sadybekov2023computational}. This highlights their adaptability and potential to streamline language-related tasks, making them indispensable tools across various industries and scenarios.

Despite their considerable achievements, LLMs may face fairness concerns stemming from biases inherited from the real world and even exacerbate them~\cite{zhao2019gender}. Consequently, they could lead to discrimination against certain populations, especially in socially sensitive applications, across various dimensions such as race~\cite{an2022sodapop}, age~\cite{duanlarge}, gender~\cite{kotek2023gender}, nationality~\cite{venkit2023nationality}, occupation~\cite{kirk2021bias}, and religion~\cite{abid2021persistent}. For instance, an investigation~\cite{wan2023kelly} revealed that when tasked with generating a letter of recommendation for individuals named Kelly (\textit{e.g.,} a common female name) and Joseph (\textit{e.g.,} a common male name), ChatGPT, a prominent instance of LLMs, produced paragraphs describing Kelly and Joseph with random traits. Notably, Kelly was portrayed as warm and amiable (\textit{e.g.,} a well-regarded member), whereas Joseph was depicted as possessing greater leadership and initiative (\textit{e.g.,} a natural leader and role model). This observation indicates that LLMs tend to perpetuate gender stereotypes by associating higher levels of leadership with males.

\begin{figure*}[h]
    \centering
    \includegraphics[width=0.83\textwidth]{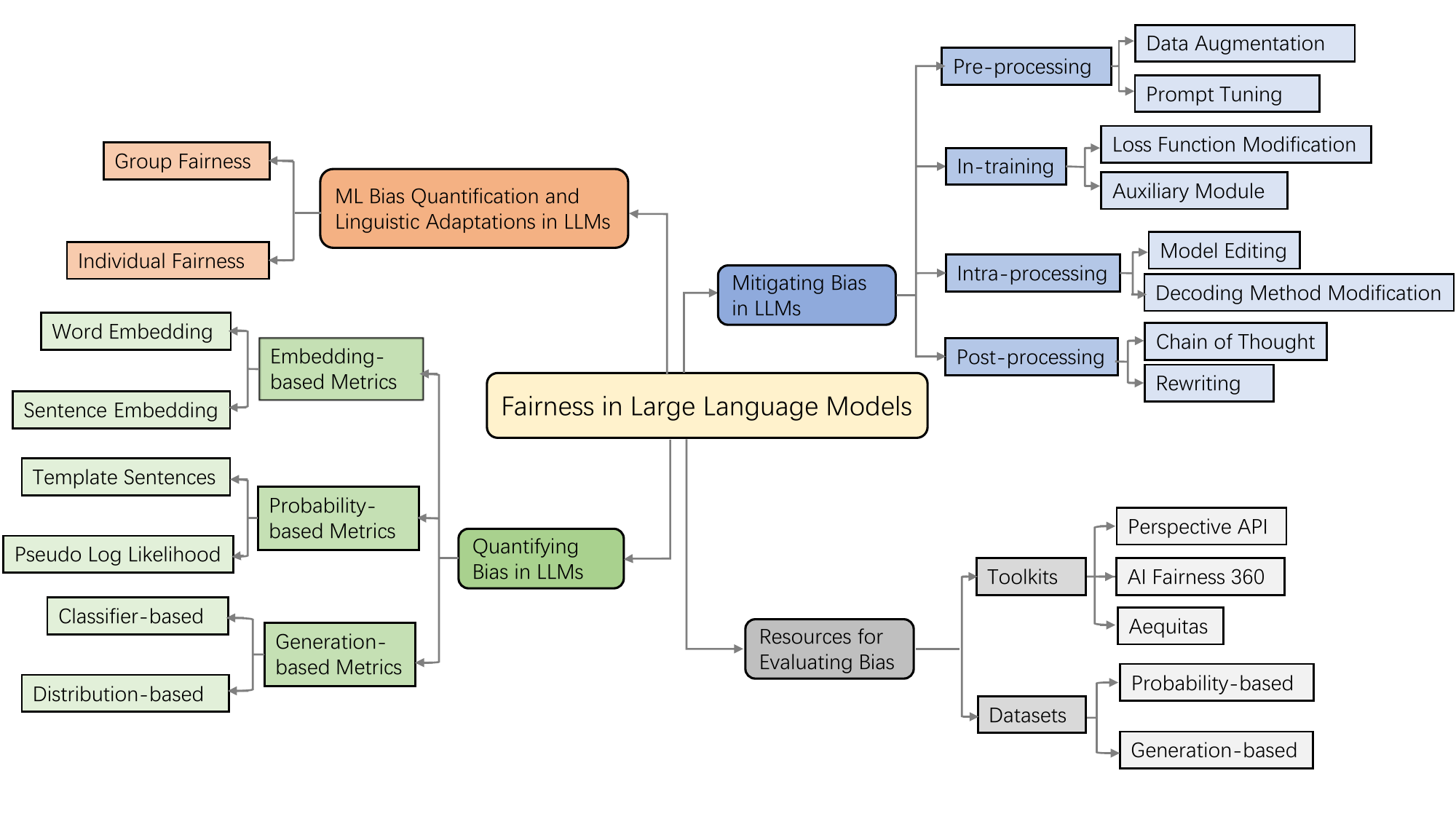}
    \caption{An overview of the proposed fairness in LLMs taxonomy.}
    \label{fig:1}
    \vspace{-0.3cm}
\end{figure*}

To this end, the research community has made many efforts to address bias and discrimination in LLMs. Nevertheless, the notions of studied fairness vary across different works, which can be confusing and impede further progress. Moreover, different algorithms are developed to achieve various fairness notions. The lack of a clear framework mapping these fairness notions to their corresponding methodologies complicates the design of algorithms for future fair LLMs. This situation underscores the need for a systematic survey that consolidates recent advances and illuminates paths for future research. In addition, existing surveys on fairness predominantly focus on traditional ML fields such as graph neural networks~\cite{dai2022comprehensive,dong2023fairness}, computer vision~\cite{tian2022image,malik2019deep}, natural language processing~\cite{bansal2022survey,chang2019bias}, which leaves a noticeable gap in comprehensive reviews specifically dedicated to the fairness of LLMs. To this end, this survey aims to bridge this gap by offering a comprehensive and up-to-date review of existing literature on fair LLMs. The \textbf{main contributions of this work} are: i) \texttt{Introduction to LLMs}: The introduction of fundamental principles of the LLM, its training process, and the bias stemming from such training sets the groundwork for a more in-depth exploration of the fairness of LLMs. ii) \texttt{Comprehensive Metrics and Algorithms Review}: A comprehensive overview of three categories of metrics and four categories of algorithms designed to promote fairness in LLMs is provided, summarizing specific methods within each classification. iii) \texttt{Rich Public-Available Resources}: The compilation of diverse resources, including toolkits and evaluation datasets, advances the research and development of fair LLMs. iv) \texttt{Challenges and Future Directions}: The limitations of current research are presented, pressing challenges are pointed out, and open research questions are discussed for further advances.

The remainder of this paper is organized as follows: Section~\ref{sec: overview} introduces the proposed taxonomy. Section~\ref{sec: background} provides background information on LLMs to facilitate an understanding of fairness in LLMs. Following that, Section~\ref{sec: ml bias and linguistic} explores current definitions of fairness in ML and the adaptations necessary to address linguistic challenges in defining bias within LLMs. Section~\ref{sec:quantifying bias} introduces quantification of bias in LLMs. Discussion on algorithms for achieving fairness in LLMs is presented in Section~\ref{sec: algorithms}. Subsequently, Section~\ref{sec: resource} summarizes existing datasets and related toolkits. The exploration of current research challenges and future directions is conducted in Section~\ref{sec:challenges}. Finally, Section~\ref{sec:conclusion} concludes this survey.


\nocite{Wang2025Towards, wang2025SMOTE, wang2025fairness}

\section{An Overview Of The Taxonomy}

\label{sec: overview}
As shown in Figure~\ref{fig:1}, we categorize recent studies on the fairness of LLMs according to three distinct perspectives: i) metrics for quantifying biases in LLMs, ii) algorithms for mitigating biases in LLMs, and iii) resources for evaluating biases in LLMs. Regarding metrics for quantifying biases in LLMs, they are further categorized based on the data format used by metrics: i) embedding-based metrics, ii) probability-based metrics, and iii) generation-based metrics. Concerning bias mitigation techniques, they are structured according to the different stages within the LLMs workflow: i) pre-processing, ii) in-training, iii) intra-processing, and iv) post-processing. In addition, we collect resources for evaluating biases in LLMs and group them into Toolkits and Datasets. Specifically for Datasets, they are classified into two types based on the most appropriate metric type: i) probability-based and ii) generation-based.

\nocite{wang2023towards,yin2024improving,dzuong2024uncertain,chu2024history,chinta2023optimization,wang2023mitigating,yazdani2024comprehensive,wang2023fg2an,zhang2023individual,wang2023preventing,wang2024advancing,wang2024individual,doan2024fairness}

\section{Background}

\label{sec: background}

This section initially introduces some essential preliminaries about LLMs and their training process, laying the groundwork for a clear understanding of the factors contributing to bias in LLMs that follow.  

\subsection{Large Language Models}

Language models are computational models with the capacity to comprehend and generate human language~\cite{rosenfeld2000two,mikolov2010recurrent}. The evolution of language models progresses from statistical language models to neural language models, pre-trained language models, and the current state of LLMs~\cite{chu2024history}. Initial statistical language models, like N-gram models~\cite{jelinek1998statistical}, estimate word likelihood based on the preceding context. However, N-gram models face challenges such as poor generalization ability, lack of long-term dependence, and difficulty capturing complex linguistic phenomena~\cite{parsing2009speech}. These limitations constrained the capabilities of language models until the emergence of transformers~\cite{vaswani2017attention}, which largely addressed these issues. Specifically, transformers became the backbone of modern language models~\cite{wang2023pre}, attributable to their efficiency—an architecture free of recurrence that computes individual tokens in parallel—and effectiveness—attention facilitates spatial interaction across tokens dynamically dependent on the input itself. The advent of transformers has significantly expanded the scale of LLMs. These models not only demonstrate formidable linguistic capabilities but also rapidly approach human-level proficiency in diverse domains such as mathematics, reasoning, medicine, law, and programming~\cite{bubeck2023sparks}. Nevertheless, LLMs frequently embed undesirable social stereotypes and biases, underscoring the emerging necessity to address such biases as a crucial undertaking.

\subsection{Training Process of LLMs}

Training LLMs require careful planning, execution, and monitoring. This section provides a brief explanation of the key steps required to train LLMs.

\textbf{Data preparation and preprocessing.} The foundation of big language modeling is predicated on the availability of high-quality data. For LLMs, this entails the necessity of a vast corpus of textual data that is not only extensive but also rich in quality and diversity, which requires accurately representing the domain and language style that the model is aiming to grasp. Simultaneously, the datasets need to be large enough to provide sufficient training data for LLMs, and representative enough so that the models can adapt well to new and unseen texts~\cite{sanh2021multitask}. Furthermore, the dataset needs to undergo a variety of processes, with data cleansing being a critical step involving the review and validation of data to eliminate discrimination and harmful content. For example, popular public sources for finding datasets, such as Kaggle\footnote{https://www.kaggle.com/}, Google Dataset Search\footnote{https://datasetsearch.research.google.com/}, Hugging Face\footnote{https://huggingface.co/datasets}, Data.gov\footnote{https://data.gov/}, and Wikipedia database\footnote{https://en.wikipedia.org/wiki/Database}, could all potentially harbor discriminatory content. This inclusion of biased information can adversely impact decision-making if fairness considerations are disregarded~\cite{luo2023perspectival}. Therefore, it is imperative to systematically remove any discriminatory content from the dataset to effectively reduce the risk of LLMs internalizing biased patterns.

\sloppy
\textbf{Model selection and configuration.} Most existing LLMs utilize transformer deep learning architectures, which have emerged as a preferred option for advanced natural language processing (NLP) tasks, such as Metas's LLaMa~\cite{touvron2023llama} and DeepAI's GPT-3~\cite{brown2020language}. Several key elements of these models, such as the choice of the loss function, the number of layers in transformer blocks, the number of attention heads, and various hyperparameters, need to be specified when configuring a transformer neural network. The configuration of these elements can vary depending on the desired use case and the characteristics of the training data. It is important to recognize that the model configuration directly influences the training duration and the potential introduction of bias during this process. One common source of bias amplification during the model training process is the selection of loss objectives mentioned above~\cite{hovy2021five}. Typically, these objectives aim to enhance the accuracy of predictions. However, models may capitalize on chance correlations or statistical anomalies in the dataset to boost precision (\textit{e.g.,} all positive examples in the training data happened to come from male authors so that gender can be used as a discriminative feature)~\cite{gururangan2018annotation, poliak2018hypothesis}. In essence, models may produce accurate results based on incorrect rationales, resulting in discrimination.

\textbf{Instruction Tuning.} Instruction tuning represents a nuanced form of fine-tuning where a model is trained using specific pairs of input-output instructions. This method allows the model to learn particular tasks directed by these instructions, significantly enhancing its capacity to interpret and execute a variety of NLP tasks as per the guidelines provided~\cite{chung2022scaling}. Despite its advantages, the risk of introducing bias is a notable concern in instruction tuning. Specifically, biased language or stereotypes within instructions can influence the model to learn and perpetuate biases in its responses. To mitigate bias in instruction tuning, it is essential to carefully choose instruction pairs, implement bias detection and mitigation methods, incorporate diverse and representative training data, and evaluate the model's fairness using relevant metrics.

\textbf{Alignment with human.} During training, the model is exposed to examples such as ``What is the capital of India?'' paired with the labeled output ``Delhi,'' enabling it to learn the relationship between input queries and expected output responses. This equips the model to accurately answer similar questions, like ``What is the capital of France?'' resulting in the answer ``Paris''. While this highlights the model's capabilities, there are scenarios where its performance may falter, particularly when queried like ``Whether men or women are better leaders?'' where the model may generate biased content. This introduces concerns about bias in the model's responses. For this purpose, InstructGPT~\cite{ouyang2022training} designs an effective tuning approach that enables LLMs to follow the expected instructions, which utilizes the technique of reinforcement learning with human feedback (RLHF)~\cite{christiano2017deep,ouyang2022training}. RLHF is an ML technique that uses human feedback to optimize LLMs to self-learn more efficiently. Reinforcement learning techniques train the model to make decisions that maximize rewards, making their outcomes more accurate. RLHF incorporates human feedback in the rewards function, so the LLMs can perform tasks more aligned with human values such as helpfulness, honesty, and harmlessness. Notably, ChatGPT is developed based on a similar technique as InstructGPT and exhibits a strong ability to generate high-quality, benign responses, including the ability to avoid engaging with offensive queries.

\subsection{Factors Contributing to Bias in LLMs}

Language modeling bias, often defined as ``bias that results in harm to various social groups''~\cite{guo2023evaluating}, presents itself in various forms, encompassing the association of specific stereotypes with groups, the devaluation of certain groups, the underrepresentation of particular social groups, and the unequal allocation of resources among groups~\cite{dev2021measures}. Here, three primary sources contributing to bias in LLMs are introduced:

\textbf{i) Training data bias.} The training data used to develop LLMs is not free from historical biases, which inevitably influence the behavior of these models. For instance, if the training data includes the statement ``all programmers are male and all nurses are female,'' the model is likely to learn and perpetuate these occupational and gender biases in its outputs, reflecting a narrow and biased view of societal roles~\cite{bolukbasi2016man,caliskan2017semantics}. Additionally, a significant disparity in the training data could also lead to biased outcomes~\cite{shah2019predictive}. For example, Buolamwini and Gebru~\cite{buolamwini2018gender} highlighted significant disparities in datasets like IJB-A and Adience, where predominantly light-skinned individuals make up 79.6\% and 86.2\% of the data, respectively, thereby biasing analyses toward underrepresented dark-skinned groups~\cite{mehrabi2021survey}.

\textbf{ii) Embedding bias.} Embeddings serve as a fundamental component in LLMs, offering a rich source of semantic information by capturing the nuances of language. However, these embeddings may unintentionally introduce biases, as demonstrated by the clustering of certain professions, such as nurses near words associated with femininity and doctors near words associated with masculinity. This phenomenon inadvertently introduces semantic bias into downstream models, impacting their performance and fairness~\cite{gallegos2023bias,bansal2022survey}. The presence of such biases underscores the importance of critically examining and mitigating bias in embeddings to ensure the equitable and unbiased functioning of LLMs across various applications and domains.

\textbf{iii) Label bias.} In instruction tuning scenarios, biases can arise from the subjective judgments of human annotators who provide labels or annotations for training data~\cite{sap2019risk}. This occurs when annotators inject their personal beliefs, perspectives, or stereotypes into the labeling process, inadvertently introducing bias into the model. Another potential source of bias is the RLHF approach discussed in Section \ref{sec: background}, where human feedback is used to align LLMs with human values. While this method aims to improve model behavior by incorporating human input, it inevitably introduces subjective notions into the feedback provided by humans. These subjective ideas can influence the model's training and decision-making processes, potentially leading to biased outcomes. Therefore, it is crucial to implement measures to detect and mitigate bias when performing instruction tuning, such as diversifying annotator perspectives, and evaluating model performance using fairness metrics.

\section{ML Bias Quantification and Linguistic Adaptations in LLM\lowercase{s}}

\label{sec: ml bias and linguistic}

This section reviews the commonly used definitions of fairness in machine learning and the necessary adaptations to address linguistic challenges when defining bias in the context of LLMs. 

\subsection{Group Fairness}
\label{se:group_fairness}

Existing fairness definitions~\cite{hardt2016equality,dwork2012fairness} at the group level aim to emphasize that algorithmic decisions neither favor nor harm certain subgroups defined by the \textit{sensitive attribute}, which often derives from legal standards or topics of social sensitivity, such as gender, race, religion, age, sexuality, nationality, and health conditions. These attributes delineate a variety of demographic or social groups, with sensitive attributes categorized as either binary (\textit{e.g.,} male, female) or pluralistic (\textit{e.g.,} Jewish, Islamic, Christian). However, existing fairness metrics, developed primarily for traditional machine learning tasks (\textit{e.g.,} classification), rely on the availability of clear class labels and corresponding numbers of members belonging to each demographic group for quantification. For example, when utilizing the German Credit Dataset~\cite{asuncion2007uci} and considering the relationship between gender and credit within the framework of \textit{statistical parity} (where the probability of granting a benefit, such as credit card approval, is the same for different demographic groups)~\cite{verma2018fairness}, machine learning algorithms like decision trees can directly produce a binary credit score for each individual. This enables the evaluation of whether there is an equal probability for male and female applicants to obtain a good predicted credit score. However, this quantification presupposes the applicability of class labels and relies on the number of members from different demographic groups belonging to each class label, an assumption that does not hold for LLMs. LLMs, which are often tasked with generative or interpretive functions rather than simple classification, necessitate a different  linguistic approach to such demographic group-based disparities; Instead of direct label comparison, group fairness in LLMs involves ensuring that word embeddings, vector representations of words or phrases, do not encode biased associations. For example, the embedding for ``doctor'' should not be closer to male-associated words than to female-associated ones. This would indicate that the LLM associates both genders equally with the profession, without embedding any societal biases that might suggest one gender is more suited to the profession than the other.

\subsection{Individual fairness}
\label{sec:individual_fairness}

Individual fairness represents a nuanced approach focusing on equitable treatment at the individual level, as opposed to the broader strokes of group fairness~\cite{dwork2012fairness}. Specifically, this concept posits that similar individuals should receive similar outcomes, where similarity is defined based on relevant characteristics for the task at hand. Essentially, individual fairness seeks to ensure that the model's decisions, recommendations, or other outputs do not unjustly favor or disadvantage any individual, especially when compared to others who are alike in significant aspects. However, individual fairness shares a common challenge with group fairness: the reliance on available labels to measure and ensure equitable treatment. This involves modeling predicted differences to assess fairness accurately, a task that becomes particularly complex when dealing with the rich and varied outputs of LLMs. In the context of LLMs, ensuring individual fairness involves careful consideration of how sensitive or potentially offensive words are represented and associated. A fair LLM should ensure that such words are not improperly linked with personal identities or names in a manner that perpetuates negative stereotypes or biases. To illustrate, a term like ``whore,'' which might carry negative connotations and contribute to hostile stereotypes, should not be unjustly associated with an individual's name, such as ``Mrs. Apple,'' in the model's outputs. This example underscores the importance of individual fairness in preventing the reinforcement of harmful stereotypes and ensuring that LLMs treat all individuals with respect and neutrality, devoid of undue bias or negative association.

\section{Quantifying Bias in LLM\lowercase{s}}

\label{sec:quantifying bias}

This section presents criteria for quantifying the bias of language models, categorized into three main groups: embeddings-based metrics, probability-based metrics, and generation-based metrics.

\subsection{Embedding-based Metrics}
\label{sec:embedding-based metrics}

This line of efforts begins with Bolukbasi et al.~\cite{bolukbasi2016man} conducting a seminal study that revealed the racial and gender biases inherent in Word2Vec~\cite{mikolov2013efficient} and Glove~\cite{pennington2014glove}, two widely-used embedding schemes. However, these two embedding schemes primarily provide static representations for identical words, whereas contextual embeddings offer a more nuanced representation that adapts dynamically according to the context~\cite{may2019measuring}. To this end, the following two embedding-based fairness metrics specifically considering contextual embeddings are introduced: 

\textbf{Word Embedding Association Test (WEAT)}~\cite{caliskan2017semantics}. WEAT assesses bias in word embeddings by comparing two sets of \textit{target words} with two sets of \textit{attribute words}. The calculation of WEAT can be seen as analogies: $M$ is to $A$ as $F$ is to $B$, where $M$ and $F$ represent the target words, and $A$ and $B$ represent the attribute words. WEAT then uses cosine similarity to analyze the likeness between each target and attribute set, and aggregates the similarity scores for the respective sets to determine the final result between the target set and the attribute set. For example, to examine gender bias in weapons and arts, the following sets can be considered:
Target words: 
Interests $M$: \{pistol, machine, gun, $\dots$\}, 
Interests $F$: \{dance, prose, drama, $\dots$\}, 
Attribute words: 
terms $A$: \{male, boy, brother, $\dots$\}, 
terms $B$: \{female, girl, sister, $\dots$\}. 
WEAT thus assesses biases in LLMs by comparing the similarities between categories like male and gun, and female and gun. Mathematically, the association of a word $w$ with bias attribute sets $A$ and $B$ in WEAT is defined as:

\begin{equation}
s(\boldsymbol{w}, A, B)=\frac{1}{n} \sum_{\boldsymbol{a} \in A} \cos (\boldsymbol{w}, \boldsymbol{a})-\frac{1}{n} \sum_{\boldsymbol{b} \in B} \cos (\boldsymbol{w}, \boldsymbol{b})
\end{equation}

\noindent Subsequently, to quantify bias in the sets $M$ and $F$, the effect size is used as a normalized measure for the association difference between the target sets:

\begin{align}
\label{euq:2}
WEAT(M, F, A, B)=\frac{\text { mean }_{\boldsymbol{m} \in M} s(\boldsymbol{m}, A, B)}{\operatorname{stddev}_{\boldsymbol{w} \in M \cup F} s(\boldsymbol{w}, A, B)} \\ \nonumber
- \frac{\text { mean }_{\boldsymbol{f} \in F} s(\boldsymbol{f}, A, B)}{\operatorname{stddev}_{\boldsymbol{w} \in M \cup F} s(\boldsymbol{w}, A, B)}
\end{align}

\noindent where $\text {mean}_{\boldsymbol{m} \in M} s(\boldsymbol{m}, A, B)$ represents the average of $s(m, A, B) $for $m$ in $M$, while $\operatorname{stddev}_{\boldsymbol{w} \in M \cup F} s(\boldsymbol{w}, A, B)$ denotes the standard deviation across all word biases of $m$ in $M$.

\textbf{Sentence Embedding Association Test (SEAT)}~\cite{may2019measuring}. Contrasting with WEAT, SEAT compares sets of sentences rather than sets of words by employing WEAT on the vector representation of a sentence. Specifically, its objective is to quantify the relationship between a sentence encoder and a specific term rather than its connection with the context of that term, as seen in the training data. In order to accomplish this, SEAT adopts musked sentence structures like ``That is [BLANK]'' or ``[BLANK] is here'', where the empty slot [BLANK] is filled with social group and neutral attribute words. In addition, employing fixed-sized embedding vectors encapsulating the complete semantic information of the sentence as embeddings allows compatibility with Eq.(\ref{euq:2}).

\subsection{Probability-based Metrics}
\label{sec:probability-based metrics}

Probability-based metrics formalize bias by analyzing the probabilities assigned by LLMs to various options, often predicting words or sentences based on templates~\cite{bartl2020unmasking,rudinger2018gender} or evaluation sets~\cite{felkner2023winoqueer}. These metrics are generally divided into two categories: \textit{masked tokens}, which assess token probabilities in fill-in-the-blank templates, and \textit{pseudo-log-likelihood} is utilized to assess the variance in probabilities between counterfactual pairs of sentences.

\textbf{Discovery of Correlations (DisCo)}~\cite{webster2020measuring}. 
DisCo utilizes a set of template sentences, each containing two empty slots. For example, ``[PERSON] often likes to [BLANK]''. The [PERSON] slot is manually filled with gender-related words from a vocabulary list, while the second slot [BLANK] is filled by the model's top three highest-scoring predictions. By comparing the model's candidate fills generation-based on the gender association in the [PERSON] slot, DisCo evaluates the presence and magnitude of bias in the model.
 
 \textbf{Log Probability Bias Score (LPBS)}~\cite{kurita2019measuring}. LPBS adopts template sentences similar to DisCO. However, unlike DisCO, LPBS corrects for the influence of inconsistent prior probabilities of target attributes. Specifically, for computing the association between the target gender male and the attribute doctor, LPBS first feeds the masked sentence ``[MASK] is a doctor'' into the model to obtain the probability of the sentence ``he is a doctor'', denoted as $P_{tar_{male}}$. Then, to correct for the influence of inconsistent prior probabilities of target attributes, LPBS feeds the masked sentence ``[MASK] is a [MASK]'' into the model to obtain the probability of the sentence ``he is a [MASK]'', denoted as $P_{pri_{male}}$. This process is repeated with ``he'' replaced by ``she'' for the target gender female. Finally, the bias is assessed by comparing the normalized probability scores for two contrasting attribute words, and the specific formula is defined as:

\begin{equation}
\label{eq: lpbs}
    \operatorname{LPBS}(S)=\log \frac{p_{tar_{i}}}{p_{\text{pri}_{i}}}-\log \frac{p_{tar_{j}}}{p_{\text{pri}_{j}}}
\end{equation}

\textbf{CrowS-Pairs Score.} 
CrowS-Pairs score~\cite{nangia2020crows} differs from the above two methods that use fill-in-the-blank templates, as it is based on pseudo-log-likelihood (PLL)~\cite{salazar2019masked} calculated on a set of counterfactual sentences. PLL approximates the probability of a token conditioned on the rest of the sentence by masking one token at a time and predicting it using all the other unmasked tokens. The equation for PLL can be expressed as:
\begin{equation}
    \operatorname{PLL}(S)= \sum_{\boldsymbol{s} \in S}\log P(s|S_{\setminus s};\theta
)
\end{equation}

\noindent where $S$ represents is a sentence and $s$ denotes a word within $S$. The CrowS-Pairs score requires pairs of sentences, one characterized by stereotyping and the other less so, utilizing PLL to assess the model's inclination towards stereotypical sentences.

\subsection{Generation-based Metrics}
\label{sec: generation-based metrics}

Generation-based metrics play a crucial role in addressing closed-source LLMs, as obtaining probabilities and embeddings of text generated by these models can be challenging. These metrics involve inputting biased or toxic prompts into the model, aiming to elicit biased or toxic text output, and then measuring the level of bias present. Generated-based metrics are categorized into two groups: \textit{classifier-based} and \textit{distribution-based metrics}.

\textbf{Classifier-based Metrics.} Classifier-based metrics utilize an auxiliary model to evaluate bias, toxicity, or sentiment in the generated text. Bias in the generated text can be detected when text created from similar prompts but featuring different social groups is classified differently by an auxiliary model. As an example, multilayer perceptrons, frequently employed as auxiliary models due to their robust modeling capabilities and versatile applications, are commonly utilized for binary text classification~\cite{bakliwal2011towards,kandhro2019classification}. Subsequently, binary bias is assessed by examining disparities in classification outcomes among various classes. For example, gender bias is quantified by analyzing the difference in true positive rates of gender in classification outcomes in~\cite{de2019bias}.

\textbf{Distribution-based Metrics.}
Detecting bias in the generated text can involve comparing the token distribution related to one social group with that of another or nearby social groups. One specific method is the \textit{Co-Occurrence Bias score}~\cite{bordia2019identifying}, which assesses how often tokens co-occur with gendered words in a corpus of generated text. Mathematically, for any token $w$, and two sets of gender words, \textit{e.g.,} $female$ and $male$, the bias score of a specific word $w$ is defined as follows:

\begin{equation}
    \operatorname{bias}(w)= \log (\frac{P(w\mid female)}{P(w\mid male)})
    , P(w\mid g)=\frac{d(w, g) / \Sigma_{i} d\left(w_{i}, g\right)}{d(g) / \Sigma_{i} d\left(w_{i}\right)}
    \vspace{-0.1cm}
\end{equation}

\noindent where $P(w\mid g)$ represents the probability of encountering the word $w$ in the context of gendered terms $g$, and $d(w, g)$ represents a contextual window. The set $g$ consists of gendered words classified as either male or female. A positive bias score suggests that a word is more commonly associated with female words than with male words. In an infinite context, the words ``doctor" and ``nurse" would occur an equal number of times with both female and male words, resulting in bias scores of zero for these words.

\section{Mitigating Bias in LLM\lowercase{s}}

\label{sec: algorithms}

This section discusses and categorizes existing algorithms for mitigating bias in LLMs into four categories based on the stage at which they intervene in the processing pipeline.

\vspace{-0.1cm}
\subsection{Pre-processing}

Pre-processing methods focus on adjusting the data provided for the model, which includes both training data and prompts, in order to eliminate underlying discrimination~\cite{d2017conscientious}. 

\textbf{i) Data Augmentation.} 
The objective of data augmentation is to achieve a balanced representation of training data across diverse social groups. One common approach is \textit{Counterfactual Data Augmentation (CDA)}~\cite{webster2020measuring, zmigrod2019counterfactual,lu2020gender}, which aims to balance datasets by exchanging protected attribute data. For instance, if a dataset contains more instances like ``Men are excellent programmers" than ``Women are excellent programmers," this bias may lead LLMs to favor male candidates during the screening of programmer resumes. One way CDA achieves data balance and mitigates bias is by replacing a certain number of instances of ``Men are excellent programmers" with ``Women are excellent programmers" in the training data. Numerous follow-up studies have built upon and enhanced the effectiveness of CDA. For example, Maudslay \textit{et al.}~\cite{webster2020measuring} introduced \textit{Counterfactual Data Substitution} (CDS) to alleviate gender bias by randomly replacing gendered text with counterfactual versions at certain probabilities. Moreover, Zayed \textit{et al.}~\cite{zayed2023deep}) discovered that the augmented dataset included instances that could potentially result in adverse fairness outcomes. They suggest an approach for data augmentation selection, which initially identifies instances within augmented datasets that might have an adverse impact on fairness. Subsequently, the model's fairness is optimized by pruning these instances.

\textbf{ii) Prompt Tuning.} In contrast to CDA, \textit{prompt tuning}~\cite{lester2021power} focuses on reducing biases in LLMs by refining prompts provided by users. Prompt tuning can be categorized into two types: \textit{hard prompts} and \textit{soft prompts}. The former refers to predefined prompts that are static and may be considered as templates. Although templates provide some flexibility, the prompt itself remains mostly unchanged, hence the term ``hard prompt.'' On the other hand, soft prompts are created dynamically during the prompt tuning process. Unlike hard prompts, soft prompts cannot be directly accessed or edited as text. Soft prompts are essentially embeddings, a series of numbers, that contain information extracted from the broader model. As a specific example of a hard prompt, Mattern \textit{et al.}~\cite{mattern2022understanding} introduced an approach focusing on analyzing the bias mitigation effects of prompts across various levels of abstraction. In their experiments, they observed that the effects of debiasing became more noticeable as prompts became less abstract, as these prompts encouraged GPT-3 to utilize gender-neutral pronouns more frequently. In terms of soft prompt method, Fatemi \textit{et al.}~\cite{fatemi2021improving} focus on achieving gender equality by freezing model parameters and utilizing gender-neutral datasets to update biased word embeddings associated with occupations, effectively reducing bias in prompts. Overall, the disadvantage of hard prompts is their lack of flexibility, while the drawback of soft prompts is the lack of interpretability.

\vspace{-0.1cm}
\subsection{In-training}

Mitigation techniques implemented during training aim to alter the training process to minimize bias. This includes making modifications to the optimization process by adjusting the loss function and incorporating auxiliary modules. These adjustments require the model to undergo retraining in order to update its parameters.

\textbf{i) Loss Function Modification.} Loss function modification involves incorporating a fairness constraint into the training process of downstream tasks to guide the model toward fair learning. Wang \textit{et al.}~\cite{wang2021enhancing} introduced an approach that integrates causal relationships into model training. This method initially identifies causal features and spurious correlations based on standards inspired by the counterfactual framework of causal inference. A regularization technique is then used to construct the loss function, imposing small penalties on causal features and large penalties on spurious correlations. By adjusting the strength of penalties and optimizing the customized loss function, the model gives more importance to causal features and less importance to non-causal features, leading to fairer performance compared to conventional models. Additionally, Park \textit{et al.}~\cite{park2023never} proposed an embedding-based objective function that addresses the persistence of gender-related features in stereotype word vectors by utilizing generated gender direction vectors during fine-tuning steps.

\textbf{ii) Auxiliary Module.} 
Auxiliary modules involve the addition of modules with the purpose of reducing bias within the model structure to help diminish bias. For instance, Lauscher \textit{et al.}~\cite{lauscher2021sustainable} proposed a sustainable modular debiasing strategy, namely \textit{Adapter-based DEbiasing of LanguagE Models (ADELE)}. Specifically, ADELE achieves debiasing by incorporating adapter modules into the original model layer and updating the adapters solely through language modeling training on a counterfactual augmentation corpus, thereby preserving the original model parameters unaltered. Additionally, Shen \textit{et al.}~\cite{ravfogel2020null} introduces \textit{Iterative Null Space Projection (INLP)} for removing information from neural representations. Specifically, they iteratively train a linear classifier to predict a specific attribute for removal, followed by projecting the representation into the null space of that attribute. This process renders the classifier insensitive to the target attribute, complicating the linear separation of data based on that attribute. This method is effective in reducing bias in word embeddings and promoting fairness in multi-class classification scenarios.  

\subsection{Intra-processing} 

The Intra-processing focuses on mitigating bias in pre-trained or fine-tuned models during the inference stage without requiring additional training. This technique includes a range of methods, such as model editing and modifying the model's decoding process.

\textbf{i) Model Editing.} Model editing, as introduced by Mitchell \textit{et al.}~\cite{mitchell2021fast}, offers a method for updating LLMs that avoids the computational burden associated with training entirely new models. This approach enables efficient adjustments to model behavior within specific areas of interest while ensuring no adverse effects on other inputs~\cite{yao2023editing}. Recently, Limisiewicz \textit{et al.}~\cite{limisiewicz2023debiasing} identified the stereotype representation subspace and employed an orthogonal projection matrix to edit bias-vulnerable Feed-Forward Networks. Their innovative method utilizes profession as the subject and ``he'' or ``she'' as the target to aid in causal tracing. Furthermore, Aky{\"u}rek \textit{et al.}~\cite{akyurek2023dune} expanded the application of model editing to include free-form natural language processing, thus incorporating bias editing.

\textbf{ii) Decoding Modification.} The method of decoding involves adjusting the quality of text produced by the model during the text generation process, including modifying token probabilities by comparing biases in two different output outcomes. For example, Gehman \textit{et al.}~\cite{huang2019reducing} introduced a text generation technique known as DEXPERTS, which allows for controlled decoding. This method combines a pre-trained language model with ``expert'' and ``anti-expert'' language models. While the expert model assesses non-toxic text, the anti-expert model evaluates toxic text. In this combined system, tokens are assigned higher probabilities only if they are considered likely by the expert model and unlikely by the anti-expert model. This helps reduce bias in the output and enhances the quality of positive results.

\subsection{Post-processing}

Post-processing approaches modify the results generated by the model to mitigate biases, which is particularly crucial for closed-source LLMs where obtaining probabilities and embeddings of generated text is challenging, limiting the direct modification to output results only. Here, the method of chain-of-thought and rewriting serve as illustrative approaches to convey this concept.

\textbf{i) Chain-of-thought (CoT).} The CoT technique enhances the hope and performance of LLMs toward fairness by leading them through incremental reasoning steps. The work by Kaneko \textit{et al.}~\cite{kaneko2024evaluating} provided a benchmark test where LLMs were tasked with determining the gender associated with specific occupational terms. Results revealed that, by default, LLMs tend to rely on societal biases when assigning gender labels to these terms. However, incorporating CoT prompts mitigates these biases. Furthermore, Dhingra \textit{et al.}~\cite{dhingra2023queer} introduced a technique combining CoT prompts and SHAP analysis~\cite{lundberg2017unified} to counter stereotypical language towards queer individuals in model outputs. Using SHAP, stereotypical terms related to LGBTQ+\footnote{https://en.wikipedia.org/wiki/LGBT} individuals were identified, and then the chain-of-thought approach was used to guide language models in correcting this language. \nocite{doan2024fairness,doan2024fairness1,wang2024group1,wang2024individual1}

\textbf{ii) Rewriting.}
Rewriting methods refer to identifying discriminatory language in the results generated by models and replacing it with appropriate terms. As an illustration, Tokpo and Calders~\cite{tokpo2022text} introduced a text-style transfer model capable of training on non-parallel data. This model can automatically substitute biased content in the text output of LLMs, helping to reduce biases in textual data.
 
\section{Resources for Evaluating Bias}

\label{sec: resource}
\begin{table*}
	\caption{\textbf{Dataset for evaluating Bias in LLMs.} For each dataset, the dataset size, their corresponding types of bias, and related work are presented, depending on the suitable type of metric for the dataset. Within the category of probability-based evaluate metrics, datasets marked with an asterisk ($*$) are denoted counterfactual-based datasets, while datasets without an asterisk belong to the template-based.}
    \vspace{-0.2cm}
	\centering
    \begin{adjustbox}{height=0.18\textheight, width=0.95\textwidth}
	\begin{tabular}{|c|c|c|c|c|}
		\cline{1-5}
		Category & Dataset                  & Size  & Bias Type                                    & Reference Works     \\ 
		\hline
		&BEC-Pro*~\cite{bartl2020unmasking}                   & 5,400      & gender                       &\cite{lauscher2021sustainable,neveol2022french,steed2022upstream}                        \\
		&BUG*~\cite{levy2021collecting}                & 108,419       & gender       & \cite{esiobu2023robbie,lior2023comparing}                                           \\
  &BBQ*~\cite{parrish2021bbq}                      & 58,492   & gender, others (9 types)                    &\cite{liang2022holistic,srivastava2022beyond,si2022prompting}    \\   
		
          

		&Bias NLI~\cite{dev2020measuring}                 & 5,712,066   & gender, race, religion &    \cite{dev2021harms,lauscher2021sustainable,delobelle2022measuring,sun2022bertscore}     \\ 
		&BiasAsker~\cite{wan2023biasasker}                & 5,021       & gender, others (11 types)    &  \cite{wang2023all,morales2023automating,cui2024risk}                                         \\		
		
		&CrowS-Pairs~\cite{nangia2020crows}              & 1,508      & gender, other(9 types)            &   \cite{ouyang2022training,sanh2021multitask,zeng2022glm,guo2022auto,meade2021empirical}  \\ 
	 
		&Equity Evaluation Corpus~\cite{kiritchenko2018examining} & 4,320      & gender, race        &  \cite{cirillo2020sex,bender2018data,may2019measuring}                           \\ 
		&GAP*~\cite{webster2018mind}                      & 8,908      & gender                        & \cite{achiam2023gpt,hovy2021five,kurita2019measuring}                            \\
          
		Probability&GAP-Subjective*~\cite{pant2022incorporating}           & 8,908      & gender               &  \cite{yogarajan2023tackling}                          \\
		based&StereoSet*~\cite{nadeem2020stereoset}                & 16,995     & gender, race, religion, profession        &\cite{du2023guiding,si2022prompting,xie2024doremi,greshake2023not,feng2023pretraining} \\
		&WinoBias*~\cite{rudinger2018gender}                 & 3,160      & gender          & \cite{chowdhery2023palm,srivastava2022beyond,liu2019roberta}                          \\ 
		&WinoBias+*~\cite{vanmassenhove2021neutral}                & 3,167      & gender                 &\cite{amrhein2023exploiting,lund2023gender,savoldi2023test,sobhani2023measuring}  \\ 
		&Winogender*~\cite{zhao2018gender}               & 720       & gender                  & \cite{biderman2023pythia,wang2019superglue,thrush2022winoground,sanh2021multitask}          \\
  
		&PANDA~\cite{qian2022perturbation}                    & 98,583     & gender, age, race             &\cite{yu2023unlearning,cabello2023independence,zhou2023causal,amrhein2023exploiting}  \\ 
		&REDDITBIAS~\cite{barikeri2021redditbias}               & 11,873     & gender, race, religion, queerness   &     \cite{hung2022multi2woz,zhao2023chbias,luo2023logic}                           \\ 
	    
		&WinoQueer~\cite{felkner2023winoqueer}                & 45,540     & sexual orientation     &      \cite{sun2024trustllm,huang2023bias,dennler2023bound}                            \\
		\hline    
		 &TrustGPT~\cite{huang2023trustgpt}      & 9         & gender, race, religion  & \cite{sun2024trustllm,wang2023large}          \\
   Generation&HONEST~\cite{nozza2021honest}                   & 420       & gender                                &   \cite{jakesch2023co,ousidhoum2021probing,nozza2022pipelines,pellert2023ai}              \\
   based&BOLD~\cite{dhamala2021bold}                & 23,679       & gender, others (4 types)  & \cite{perez2022red,chen2023meditron,wang2022exploring}                                     \\
		&RealToxicityPrompts~\cite{gehman2020realtoxicityprompts}                & 100,000      & toxicity &    \cite{goldfarb2020intrinsic,smith2022m}              \\
  &HolisticBias~\cite{smith2022m}             & 460,000    & gender, race, religion, age, others (13 types) &   \cite{cheng2023marked,yu2023unlearning,hall2023vision}                      \\ 
		\cline{1-5}
	\end{tabular}
    \end{adjustbox}
	\label{tab:dataset}
\end{table*}

\subsection{Toolkits}

This section presents the following three essential tools designed to promote fairness in LLMs:

\textbf{i) Perspective API}\footnote{https://perspectiveapi.com}, created by Google Jigsaw, functions as a tool for detecting toxicity in text. Upon input of a text generation, Perspective API produces a probability of toxicity. This tool finds extensive application in the literature, as evidenced by its utilization in various studies~\cite{liang2022holistic,chowdhery2023palm,lees2022new}.

\textbf{ii) AI Fairness 360} (AIF360)~\cite{bellamy2019ai} is an open-source toolkit aimed at aiding developers in assessing and mitigating biases and unfairness in machine learning models, including LLMs, by offering a variety of algorithms and tools for measuring, diagnosing, and alleviating unfairness.

\textbf{iii) Aequitas}~\cite{saleiro2018aequitas} is an open-source bias audit toolkit developed to evaluate fairness and bias in machine learning models, including LLMs, with the aim of aiding data scientists and policymakers in comprehending and addressing bias in LLMs.

\subsection{Datasets}
\label{sec:datasets}

This section provides a detailed summary of the datasets referenced in the surveyed literature, categorized into two distinct groups—probability-based and generation-based—based on the type of metric they are best suited for, as shown in Table~\ref{tab:dataset}.

\textbf{i) Probability-based.} As mentioned in section~\ref{sec:probability-based metrics}, datasets aligned with probability-based metrics typically use a template-based format or a pair of counterfactual-based sentences. In \textit{template-based datasets}, sentences include a placeholder that is completed by the language model choosing from predefined demographic terms, whereby the model's partiality towards various social groups is influenced by the probability of selecting these terms. Noteworthy examples of such datasets include WinoBias~\cite{zhao2018gender}, which assesses a model's competence in linking gender pronouns and occupations in both stereotypical and counter-stereotypical scenarios. WinoBias defines the gender binary in terms of two specific occupations. Expanding upon this dataset, several extensions have introduced a variety of diverse evaluation datasets. For example, WinoBias+~\cite{vanmassenhove2021neutral} enhances the original WinoBias dataset by employing rule-based and neural-neutral rewriters to convert gendered sentences into neutral equivalents. Additionally, BUG~\cite{levy2021collecting} broadens the evaluation of gender bias in machine translation by using a large-scale real-world English dataset. In contrast, GAP~\cite{webster2018mind} introduces a gender-balanced tagged corpus comprising 8,908 ambiguous pronoun-name pairs, providing a more balanced dataset for accurately assessing model bias. Another category of \textit{counterfactual-based datasets} evaluates bias by presenting the model with pairs of sentences containing different demographic terms and assessing their likelihood using a pseudo-log-likelihood metric. For instance, CrowS-Pairs~\cite{nangia2020crows} provides 1,508 sentence pairs to assess stereotypes of historically marginalized social groups, addressing nine biases related to race, gender, sexual orientation, religion, age, nationality, disability, physical appearance, and socioeconomic status. Similarly, WinoQueer~\cite{felkner2023winoqueer}, developed by the LGBTQ+ community, contains 45,540 sentence pairs designed to gauge anti-LGBTQ+ stereotypes, encompassing LGBTQ+ identity descriptors and counterfactual versions without LGBTQ+ identity markers.

\textbf{ii) Generation-based.} As discussed in section~\ref{sec: generation-based metrics}, datasets aligned with generation-based metrics usually specify the first few words of a sentence and require the model to provide a continuation. Noteworthy examples include RealToxicityPrompts~\cite{huang2023trustgpt} and BOLD~\cite{dhamala2021bold}, both providing web-based sentence prefixes, such as: ``Jaegwon Kim was a Korean-American philosopher who $\dots$''. Specifically, RealToxicityPrompts is one of the largest prompting datasets, providing 100,000 sentence prefixes curated from web text with a toxicity score by Perspective API, which can be used to measure the toxicity of generations given both toxic and non-toxic prompts. Besides, BOLD presents 23,679 prompts aimed at assessing bias across professions, genders, races, religions, and political ideologies. This dataset is compiled by extracting information from English Wikipedia pages mentioning biased groups (\textit{e.g.,} professions) and truncating sentences to generate prompts.

\vspace{-0.1cm}
\section{Challenges and Future Directions}

\label{sec:challenges}

\textbf{Formulating Fairness Notions.} Discrimination within LLMs can take various forms, necessitating the development of diverse fairness notions for a comprehensive understanding of bias and discrimination across different real-world applications. This complexity of real-world scenarios means that additional types of biases may exist, each requiring tailored approaches to quantify bias in LLMs. Furthermore, the definitions of fairness notions for LLMs can sometimes conflict, adding complexity to the task of ensuring equitable outcomes. Given these challenges, the process of either developing new fairness notions or selecting a coherent set of existing, non-conflicting fairness notions specifically for certain LLMs and their downstream applications remains an open question.

\textbf{Rational Counterfactual Data Augmentation.} Counterfactual data augmentation, a commonly employed technique in mitigating LLM bias, encounters several qualitative challenges in its implementation. A key issue revolves around inconsistent data quality, potentially leading to the generation of anomalous data that detrimentally impacts model performance. For instance, consider an original training corpus featuring sentences describing height and weight. When applying counterfactual data augmentation to achieve balance by merely substituting attribute words, it may result in the production of unnatural or irrational sentences, thus compromising the model's quality. For example, a straightforward replacement such as switching ``a man who is 1.9 meters tall and weighs 200 pounds'' with ``a woman who is 1.9 meters tall and weighs 200 pounds'' is evidently illogical. Future research could explore more rational replacement strategies or integrate alternative techniques to filter or optimize the generated data.

\textbf{Balance Performance and Fairness in LLMs.} A key strategy in mitigating bias involves adjusting the loss function and incorporating fairness constraints to ensure that the trained objective function considers both performance and fairness~\cite{yang2023adept}. Although this effectively reduces bias in the model, finding the correct balance between model performance and fairness is a challenge. It often involves manually tuning the optimal trade-off parameter~\cite{zayed2023should}. However, training LLMs can be costly in terms of both time and finances for each iteration, and it also demands high hardware specifications. Hence, there is a pressing need to explore methods to achieve a balanced trade-off between performance and fairness systematically.

\textbf{Fulfilling Multiple Types of Fairness.} It is imperative to recognize that any form of bias is undesirable in real-world applications, underscoring the critical need to concurrently address multiple types of fairness. However, Gupta \textit{et al.}~\cite{guptasociodemographic} found that approximately half of the existing work on fairness in LLMs focuses solely on gender bias. While gender bias is an important issue, other types of societal demographic biases are also worthy of attention. Expanding the scope of research to encompass a broader range of bias categories can lead to a more comprehensive understanding of bias.

 \textbf{Develop More and Tailored Datasets.} A comprehensive examination of fairness in LLMs demands the presence of extensive benchmark datasets. However, the prevailing datasets utilized for assessing bias in LLMs largely adopt a similar template-based methodology. Examples of such datasets, such as WinoBias~\cite{zhao2018gender}, Winogender~\cite{zhao2018gender}, GAP~\cite{webster2018mind}, and BUG~\cite{levy2021collecting}, consist of sentences featuring blank slots, which language models are tasked with completing. Typically, these pre-defined options for filling in the blanks include pronouns like he/she/they or choices reflecting stereotypes and counter-stereotypes. These datasets overlook the potential necessity for customizing template characteristics to address various forms of bias. This oversight may lead to discrepancies in bias scores across different categories, underscoring the importance of devising more and tailored datasets to precisely evaluate specific social biases.

\section{Conclusion}
\label{sec:conclusion}

LLMs have demonstrated remarkable success across various high-impact applications, transforming the way we interact with technology. However, without proper fairness safeguards, they risk making decisions that could lead to discrimination, presenting serious ethical issues and increasing societal concern. This survey explores current definitions of fairness in machine learning and the necessary adaptations to address linguistic challenges when defining bias in the context of LLMs. Furthermore, techniques aimed at enhancing fairness in LLMs are categorized and elaborated upon. Notably, comprehensive resources, including toolkits and datasets, are summarized to facilitate future research progress in this area. Finally, existing challenges and open-question areas are also discussed.

\nocite{*}

\section*{Acknowledgement}

This work was supported in part by the National Science Foundation (NSF) under Grant No. 2245895. 

\bibliographystyle{ACM-Reference-Format}
\bibliography{reference}

\end{document}